\documentclass[11pt,table]{article}

\usepackage[final]{acl2023}

\usepackage{times}
\usepackage{latexsym}
\usepackage{graphicx}
\usepackage{amsmath,amssymb,amsfonts}
\usepackage{multirow}
\usepackage{soul}
\usepackage{booktabs}
\usepackage{threeparttable}
\usepackage{float}
\usepackage{colortbl}
\usepackage{subcaption}
\usepackage{makecell}
\usepackage[T1]{fontenc}
\usepackage[utf8]{inputenc}
\usepackage{microtype}
\usepackage{inconsolata}

\title{Improving French Synthetic Speech Quality via SSML Prosody Control}

\author{
  Nassima Ould Ouali$^{1}$,
  Awais Hussain Sani$^{2}$,
  Ruben Bueno$^{1}$\footnotemark[2], \\
  \textbf{Jonah Dauvet}$^{1,3}$\footnotemark[2]\textbf{,}
  \textbf{Tim Luka Horstmann}$^{1,2}$\footnotemark[2]\textbf{,}
  \textbf{Eric Moulines}$^{1}$ \\
  $^{1}$École Polytechnique, France,\;
  $^{2}$Hi! PARIS Research Center, France,\\
  $^{3}$McGill University, Canada \\
  {\small \texttt{\{nassima.ould-ouali, ruben.bueno, eric.moulines\}@polytechnique.edu}} \\
  {\small \texttt{\{awais.sani, tim.horstmann\}@ip-paris.fr, jonah.dauvet@mail.mcgill.ca}} \\
}

\begin{document}
\maketitle
\begingroup
  \renewcommand\thefootnote{\fnsymbol{footnote}} % 1=*, 2=†, 3=‡, ...
  \footnotetext[2]{Equal contribution; authors listed in alphabetical order.}
\endgroup

\begin{abstract}

Despite recent advances, synthetic voices often lack expressiveness due to limited prosody control in commercial text-to-speech (TTS) systems. We introduce the first end-to-end pipeline that inserts Speech Synthesis Markup Language (SSML) tags into French text to control pitch, speaking rate, volume and pause duration. We employ a cascaded architecture with two QLoRA-fine-tuned Qwen 2.5-7B models: one predicts phrase-break positions and the other performs regression on prosodic targets, generating commercial TTS-compatible SSML markup. Evaluated on a 14-hour French podcast corpus, our method achieves 99.2\% $F_1$ for break placement and reduces mean absolute error on pitch, rate, and volume by 25–40\% compared with prompting-only large language models (LLMs) and a BiLSTM baseline. In perceptual evaluation involving 18 participants across over 9 hours of synthesized audio, SSML-enhanced speech generated by our pipeline significantly improves naturalness, with the mean opinion score increasing from 3.20 to 3.87 ($p<0.005$). Additionally, 15 of 18 listeners preferred our enhanced synthesis. These results demonstrate substantial progress in bridging the expressiveness gap between synthetic and natural French speech. Our code is publicly available at \url{https://github.com/hi-paris/Prosody-Control-French-TTS}.
\end{abstract}

\section{Introduction}
\label{sec:Intro}
Recent Text-to-Speech (TTS) advances have improved speech intelligibility; yet, achieving natural and expressive prosody remains challenging. Commercial TTS solutions prioritize clarity over prosodic variation, resulting in a monotonous speech output. This limitation particularly affects French due to its complex prosodic features.

\textbf{Speech Synthesis Markup Language (SSML)} provides a standardized way to control prosodic features such as pitch, speaking rate, and volume. Unlike neural models, SSML allows post-hoc adjustments and is compatible with commercial TTS engines. Yet, automating SSML generation is difficult: manual markup does not scale, and current LLM-based methods often produce incomplete tags, invalid syntax, or imprecise prosodic control.

We propose an automated SSML pipeline for French, combining structured prosody extraction with a novel cascaded LLM approach for simultaneous tag prediction and prosodic parameter regression. Key contributions include: \begin{itemize}
\item \textbf{End-to-end SSML annotation pipeline} that aligns speech to text, segments input into prosodic syntagms, and extracts prosodic coefficients normalized relative to a commercial TTS baseline. 
\item \textbf{Rigorous benchmarking} comparing state-of-the-art (SOTA) approaches (fine-tuned BERT, BiLSTM) with contemporary LLMs across varied prompting strategies and metrics.
\item \textbf{Cascaded LLM architecture} using two fine-tuned Qwen 2.5-7B models: one for SSML structure/boundaries and another for prosodic prediction, ensuring valid markup and accurate parameter control.
\end{itemize}

\section{Related Work} 
\label{sec:relatedWork}
Enhancing neural TTS prosody through automatic markup is an active research domain categorized into: (i) \textit{learning paradigm} (supervised vs. unsupervised approaches) and (ii) \textit{prosodic objective} (prominence, phrasing, style).

\paragraph{Supervised Prosody Learning\\}

\textbf{Word-level prominence modeling emphasizes salient words using prosodic cues} like pitch and duration. \citet{stephenson_bert_2022} fine-tune BERT~\cite{devlin_bert_2019} to predict three-level prominence tags from wavelet-based labels, achieving $F_1 = 0.588$ and enabling controllable synthesis in FastSpeech~2. Similarly, \citet{zhong2023ee} integrate emphasis features into FastSpeech~2, improving expressiveness (+0.49 Mean Opinion Score (MOS)) and naturalness (+0.67 MOS).

\textbf{Prosodic emphasis prediction controls automated stress placement patterns.} \citet{shechtman2021supervised} employ a hybrid model with acoustic and syntactic features, and \citet{seshadri2021emphasis} propose a hierarchical latent model. \citet{liu2024emphasis} combine graph-based contextual encoding with FastSpeech~2 for enhanced rendering. More recently, \citet{chen2025drawspeech} present DrawSpeech, a user-sketched prosodic contour control.

\textbf{Phrasing segments speech into natural prosodic units with appropriate pauses.} Transformer-based models now outperform recurrent neural networks (RNNs) for break prediction: \citet{futamata2021phrase} integrate BERT embeddings with linguistic features, improving phrase break prediction ($F_1$ +3.2 points, MOS = 4.39). \citet{vadapalli2025investigation} show that fine-tuned BERT outperforms RNN baselines, reaching $F_1 = 0.92$ and achieving 58.5\% listener preference for BERT-guided punctuation in narrative TTS.

\textbf{LLMs enable automated emotional and stylistic annotations at scale}. \citet{yoon2022language} prompt GPT-3 to assign sentence-level emotion labels that guide expressive TTS, achieving MOS 3.92 (naturalness) and 3.94 (expressiveness), matching human-annotated systems. Complementarily, \citet{burkhardt2023going} show that even simple, rule-based SSML adaptations can shape emotional perception, with Unweighted Average Recall scores of 0.76 for arousal and 0.43 for valence.% two key dimensions of affective speech.

\textbf{Narrative prosody modeling adjusts pitch, speaking rate, and volume to enhance expressive storytelling.} \citet{pethe2025prosodyanalysisaudiobooks} use MPNet embeddings and BiLSTMs to predict phrase-level prosody from text. Their SSML-integrated predictions improved alignment with human narration in 22–23 out of 24 audiobooks, yielding +50\% listener preference over commercial baselines.

\paragraph{Unsupervised Prosody Learning\\}

\textbf{Discrete prosody representations eliminate dependency on manual annotations} by learning prosodic patterns directly from speech data. \citet{korotkova2024word} utilize a vector-quantized variational autoencoder with Wav2Vec2 and RoBERTa encodings, deriving ten interpretable prosodic tags that enhance TTS expressiveness across multiple languages,  confirmed by MOS tests $(p<0.001)$.
In contrast, \citet{karlapati2021prosodic} learn continuous 64-dimensional prosody embeddings: a VAE encodes mel-spectrograms, and a RoBERTa + syntax-GNN regresses these from text. At inference, the 64-dimensional prosodic code conditions a Tacotron2 decoder, yielding 13.2\% comparative MOS gain (3.30→3.74, $p<0.005$) on LJSpeech with $F_0$ correlation of $r=0.68$. Discrete tags offer interpretability; continuous embeddings better capture fine-grained intonation. Both improve TTS expressiveness without hand-crafted annotations.

\paragraph{Limitations of Prior Work:}

However, existing research exhibits critical gaps. Current methods lack a comprehensive end-to-end framework for converting raw speech into standardized SSML-compliant prosodic markup. Most rely on partial manual annotations, address isolated prosodic control aspects, or produce markup incompatible with commercial TTS systems. Furthermore, the majority of existing work focuses on English, leaving other complex languages like French underexplored.
Additionally, current LLM-based approaches suffer from systematic limitations, undergenerating necessary tags, producing syntactically invalid SSML structures, and lacking precise control over numerical prosodic parameters, which prevents deployment in practical TTS systems.

We address these limitations with two main contributions: (i) we introduce the first reproducible, comprehensive French pipeline that automatically extracts fine-grained prosodic annotations and converts them into standards-compliant SSML, and (ii) we develop a novel cascaded LLM architecture that generates syntactically correct prosodic tags with precise numerical control at inference time, resulting in substantially enhanced naturalness and expressiveness in synthetic speech.

\section{Dataset Creation and SSML Annotation Pipeline}
\label{sec:dataset}

\begin{figure*}[htbp!]
    \centering
    \includegraphics[width=1\linewidth]{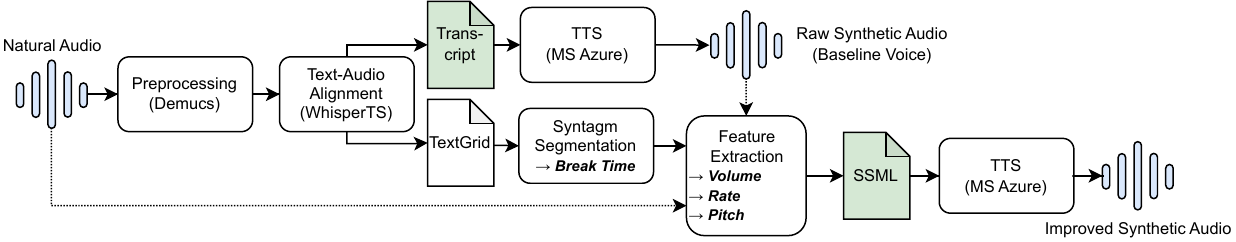}
    \caption{Overview of the SSML annotation pipeline. Natural speech is aligned, segmented, and compared to a synthetic baseline to extract prosodic features for SSML markup. Green elements indicate later model training data.}
    \label{fig:pipeline}
\end{figure*}
We construct a comprehensive dataset annotated with prosodic features from French speech. Our methodology involves aligning spoken audio with transcripts, extracting four key prosodic features --- \textbf{pitch}, \textbf{volume}, \textbf{speaking rate}, and \textbf{break duration} --- and converting them into standardized SSML for enhanced synthetic speech generation. Figure~\ref{fig:pipeline} presents our preprocessing pipeline. Further dataset statistics are provided in Appendix~\ref{sec:appendix_dataset}.

\paragraph{\textbf{Audio Collection and Preprocessing:}}
\label{sec:audio_collection}

We process 14 hours of diverse French audio content sourced from \textit{ETX Majelan}\footnote{\url{https://etxmajelan.com/}}, a high-quality podcast platform with interviews and discussions. Our dataset includes speech from 14 distinct speakers (42\% female). The original recordings contain background music, jingles, and sound effects, complicating prosodic analysis.
Hence, we isolate clean speech using \textit{Demucs}~\cite{defossez2021hybrid}, a SOTA audio source separation model, down-sample to 16 kHz, and peak-normalize the audio. Using \textit{pydub}\footnote{\url{https://github.com/jiaaro/pydub}}, we segment the cleaned audio via silence detection with a $-35$~dBFS threshold and 300~ms gaps. The resulting audio segments serve as our fundamental processing units for subsequent prosodic analysis.

\paragraph{\textbf{Text-Audio Alignment:}}  
\label{sec:text_audio_alignment}
Accurate alignment between audio and transcribed text is crucial for prosody extraction, but particularly challenging in French due to phonetic phenomena such as liaisons, elisions, and prosodic contractions. To address this, we employ the Whisper Timestamped package\footnote{\url{https://github.com/linto-ai/whisper-timestamped}} with the Whisper~\cite{radford2022robust} medium model and Auditok Voice Activity Detection (VAD)~\footnote{\url{https://github.com/amsehili/auditok}}, which filters out silent segments that would otherwise distort prosodic measurements.

To evaluate this setup, we benchmarked it against larger Whisper models, Montreal Forced Aligner (MFA) by ~\citet{mcauliffe_montreal_2017}, and NVIDIA NeMo ~\cite{kuchaiev_nemo_2019}. Benchmarking used our dataset and FLEURS benchmark \cite{conneau2022fleursfewshotlearningevaluation}  as a state-of-the-art reference.. While larger models yielded marginal gains, they introduced instability such as significant hallucinations during silence -- a known issue in Whisper~\cite{baranski_investigation_2025} -- as well as higher computational cost. Our chosen configuration achieved a 5.95\% WER using Whisper-medium with an average Alignment Recall Rate (ARR) of 96.3\% over 15-second windows against the manual TextGrid annotations created with Praat~\cite{boersma2001speak} (see Table~\ref{tab:eval_whisper}). This yielded an optimal accuracy-efficiency trade-off.

\begin{table*}[htbp!]
  \centering
  \caption{Metric evaluation of four whisper models, MFA, and NeMo on our dataset (Section~\ref{sec:dataset}) and \textcolor{blue}{FLEURS}.}
  \label{tab:eval_whisper}
  \vspace{0.5ex}
  \begin{threeparttable}
    \footnotesize
    \resizebox{1\textwidth}{!}{%
    \begin{tabular}{lcccccc}
      \toprule
        & \multicolumn{4}{c}{Whisper Model Variants} & \multicolumn{2}{c}{Alignment Models} \\
      \cmidrule(lr){2-5} \cmidrule(lr){6-7}
      Metric
        & Medium
        & Large v2
        & Large v3
        & Turbo Large v3
        & MFA $\ddagger$ 
        & NeMo Large $\ddagger$  \\
      \midrule
      Parameters
        & 769 M
        & 1550 M
        & 1540 M
        & 809 M
        & –
        & 120 M  \\
      WER$^\dagger$ 
        & 5.95\% / \textcolor{blue}{10.70\%}
        & 4.60\% / \textcolor{blue}{6.27\%}
        & 3.92\% / \textcolor{blue}{5.80\%}
        & 3.52\% / \textcolor{blue}{5.71\%}
        & –
        & –  \\
      WER$^\dagger$ +VAD
        & 5.68\% / \textcolor{blue}{8.72\%}
        & 5.07\% / \textcolor{blue}{6.16\%}
        & 3.86\% / \textcolor{blue}{5.65\%}
        & 6.16\% / \textcolor{blue}{5.83\%}
        & –
        & –  \\
      ARR$^*$ 
        & 96.3\%
        & 97.1\%
        & 96.2\%
        & 97.8\%
        & 99.7\%
        & 50.7\%   \\
      Start MAE$^*$  (ms)
        & 264
        & 191
        & 207
        & 152
        & 115
        & 4529  \\
      Duration MAE$^*$  (ms)
        & 91
        & 77
        & 102
        & 76
        & 95
        & 218  \\
      \bottomrule
    \end{tabular}
    }%

    \vspace{0.5ex}
    \begin{tablenotes}
      \scriptsize
      \item$\dagger$  WER computed with the HuggingFace \texttt{evaluate} library.
      \item$*$  ARR and MAE are computed on 15 second segments, against the gold manually annotated Text Grids of 1 hour of speech from our dataset.
      \item$\ddagger$ MFA and NeMo alignments use gold transcripts, thus rendering the WER 0 by default.
    \end{tablenotes}
  \end{threeparttable}
\end{table*}

% $WER = \frac{S + D + I}{N}$
      % Where:
      % S = Number of substitutions
      % D = Number of deletions
      % I = Number of insertions
      % N = Number of words in the reference (gold) transcript
      %
    
% MAE (Mean Absolute Error):
      % $MAE = \frac{1}{n} \sum_{i=1}^{n} |P_i - A_i|$
      % Where:
      % n = Number of predictions
      % P_i = Predicted value for the i-th instance
      % A_i = Actual (gold) value for the i-th instance
      %
% ARR (Alignment Recall Rate): This is often custom, but generally:
      % $ARR = \frac{\text{Number of correctly aligned words}}{\text{Total number of words in reference}}$
      % Or sometimes defined more strictly with tolerance:
      % $ARR = \frac{\text{Number of aligned words within tolerance}}{\text{Total number of words in reference}}$

\paragraph{\textbf{Baseline Voice for Prosodic Comparison:}}

For prosodic reference, we synthesize each transcript using \texttt{Microsoft Azure Neural TTS} with the French voice \textit{Henri}~\cite{microsoft2024ssml}. Henri was selected for its clarity, broad phonetic coverage, and consistent yet neutral prosodic characteristics, making it optimal for computing relative prosodic adjustments. The resulting synthetic speech provides a stable baseline against which natural prosodic features are measured and compared, as detailed in subsequent sections.

\paragraph{\textbf{Syntagm Segmentation:}}

Each segment undergoes further subdivision into \textit{syntagms}: prosodic units with natural pause boundaries. Following ~\citet{roll2023psst}, we detect them through acoustic pauses and punctuation.
We first derive a word/pause sequence from the TextGrid, where pauses following function words are discarded with a POS filter to remove Whisper artifacts. Next, any silence that follows \textit{.}, \textit{?}, or \textit{!} is clamped to at least 500 ms, and a 500 ms pause is injected whenever Whisper failed to signal the end of a sentence. The resulting timestamped syntagms provide stable, linguistically meaningful units for prosodic analysis.

\paragraph{\textbf{Prosodic Feature Extraction and SSML Tag Construction:}}
Each syntagm is annotated with four prosodic features: median \textbf{pitch} (fundamental frequency $f_0$), segment-level \textbf{volume} (Loudness Units Full Scale (LUFS)), \textbf{speaking rate} (words per second), and inter-syntagmatic \textbf{break duration}. All features are computed for both natural and synthetic baseline voices to derive relative delta values for SSML encoding.
To account for intra- and inter-speaker variability, we normalize each syntagm’s pitch, volume, and rate relative to a baseline computed as the median over a sliding window of $w=10$ audio segments (or, when $w$ covers all segments, the global median). The computation of each feature is detailed as follows:

\textbf{Pitch} median fundamental frequency $f_0^{(i)}$ is converted to a semitone offset  
$ s_i = 12 \log_2\!\bigl( f_0^{(i)} / \bar f_0 \bigr)$,  
clipped to $\bigl[-0.7P,\;P\bigr]$ to allow larger upward than downward shifts, and re-scaled to percentage pitch change  
$ p_i = \bigl(2^{s_i/12}-1\bigr)\!\times\!100$.

Using LUFS for \textbf{volume}, the baseline–synthetic difference 
$\Delta L_i = \bar L - L_{\mathrm{syn}}^{(i)}$  
is mapped to a gain  
$ v_i = \bigl(10^{\Delta L_i/20}-1\bigr)\!\times\!100$,  
then clipped to $\pm V$ (we use $V=10\%$).

\textbf{Speaking rate} is estimated as \textit{words per second}.  
Let $n_i$ be the word count and $d_{\mathrm{nat}}$, $d_{\mathrm{syn}}$ the net speaking durations (pauses removed).  
The rate delta is  
$r_i = \frac{n_i/d_{\mathrm{nat}} - n_i/d_{\mathrm{syn}}}{n_i/d_{\mathrm{syn}}}\!\times\!100$.  
Slow-downs are amplified for long syntagms ($>1$ s) while speed-ups are reduced, and the final value is clamped to $\pm R$ with a tighter \mbox{$+0.5R$} ceiling for accelerations.

To improve prosodic smoothness, we apply exponential smoothing to pitch and rate with $\alpha = 0.2$:
\[
\tilde{x}_0 = x_0, \quad \tilde{x}_i = \alpha x_i + (1 - \alpha) \tilde{x}_{i-1}.
\]
Sudden jumps are clamped to $\Delta = 8\%$ per syntagm. Volume is not smoothed.

\textbf{Break durations} are taken from the inter-syntagm silence gaps and inserted as raw durations (e.g., \texttt{<break time="200ms">}). The final SSML markup is assembled by inserting appropriate \texttt{<prosody>} and \texttt{<break>} tags into the text.\footnote{At inference, we found that wrapping each \texttt{<prosody>} tag with \texttt{<mstts:si\-lence type="leading-exact/trailing-exact" value="0"/>} improves output by suppressing unwanted Azure TTS pauses.}

\section{Methodology}
\label{sec:methodology}

We test whether text alone encodes sufficient cues for prosody by training two baselines: (i) a BERT-base model fine-tuned for token-level pause prediction \cite{vadapalli2025investigation}, and (ii) a BiLSTM \cite{pethe2025prosodyanalysisaudiobooks} which predicts SSML tags with pitch, speaking rate and volume adjustments.

\subsection{Fine-tuning BERT for Pause Prediction}
\label{subsec:bert_baseline}

Following \citet{vadapalli2025investigation}, we fine-tune an uncased BERT-base model for token-level pause prediction. A binary classification head determines whether each sub-word is followed by a break tag. We adopt the same hyperparameters as the original work on our dataset: batch size 64, learning rate $10^{-5}$, and gradient clipping at 10. For evaluation, we report both $F_1$ score and perplexity. While $F_1$ is used in \citet{vadapalli2025investigation}, perplexity is introduced here as an additional metric to enable broader comparisons in later sections.

We additionally introduce bootstrapping, a technique not used in the original paper, to evaluate the small model's variance in performance. We bootstrap on 10 distinct sets with the same configuration as the original training set, which allows us to obtain a distribution of performance scores for robust estimation of the uncertainty of performance. Given the reduced size of the dataset, we expect overall performance to degrade slightly. Hence, we focus on stability, measured via standard deviation.

\subsection{BiLSTM-Based Sequence Modeling}
\label{subsec:bilstm_baseline}

We implement a BiLSTM baseline following \citet{pethe2025prosodyanalysisaudiobooks}, explicitly modeling prosody prediction as a sequence regression task to predict three SSML parameters: pitch, volume, and speaking rate. This approach leverages local context through sequential processing of prosodic units.

Each syntagm receives encoding into a 768-dimensional representation using the pretrained sentence encoder \texttt{all-mpnet-base-v2}\footnote{\url{https://huggingface.co/sentence-transformers/all-mpnet-base-v2}}. We construct overlapping input sequences of varying lengths $L \in \{1, 2, 3, 4\}$, extending beyond the original study's sequence lengths of 2 and 3 to assess optimal context window size. targeting z-scored prosody vectors (pitch, volume, rate) of central segments, normalized on training statistics.

The architecture includes LayerNorm preprocessing, bidirectional LSTM (40 units per direction), dense layer (20 units, $\tanh$ activation), and linear projection for predicting the 3-dimensional prosody vector. Training uses MSE loss between predicted and target z-scored vectors. We additionally compute raw RMSE and MAE metrics for interpretability and literature comparison.

\subsection{Zero-shot and Few-shot Evaluation}
\label{subsect:FS_LLMs}

To assess SOTA LLMs for SSML markup generation, we benchmarked various open-source models in both \textbf{zero-shot} and \textbf{few-shot} settings. We evaluated Mistral (7B), Qwen 2.5 (7B), Llama 3 (8B), Granite 3.3 (8B), Qwen 3 (8B), DeepSeek-R1 (32B), and Qwen 3 (32B) via the Ollama framework\footnote{\url{https://ollama.com/}}. Models were prompted at the segment level ($\approx$ eight tags per segment on average) with French text, and tasked with generating fully annotated SSML for 100 randomly chosen segments. Few-shot prompts included 10 reference examples.

\subsection{Cascaded Fine-tuning Approach}
\label{subsec:cascaded_finetune}

\begin{figure*}[htbp!]
    \centering
    \includegraphics[width=1\linewidth]{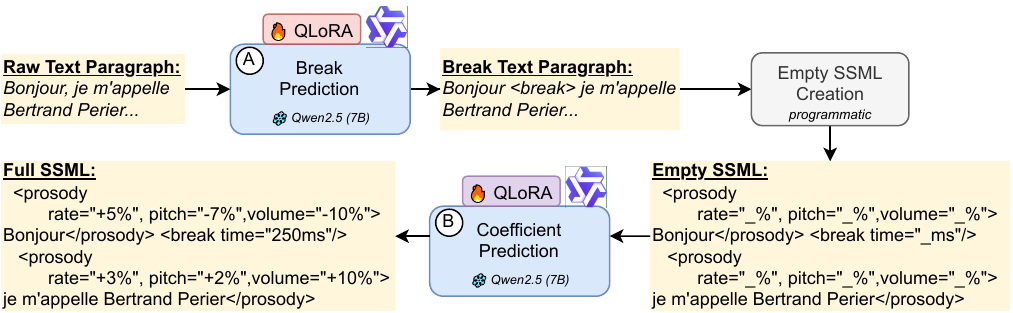}
    \caption{Cascaded LLM approach for automated text-to-SSML generation: QwenA predicts tag placement, QwenB injects prosodic values. This disentangled design enables accurate and efficient prosody control for synthetic speech.}
    \label{fig:cascade}
\end{figure*}

As we show in Section~\ref{subsec:prompting_results}, LLM-based approaches under-generate \texttt{<break>} and \texttt{<prosody>} tags, resulting in SSML that is structurally incomplete and limited in expressive control. To address this, we introduce a cascaded strategy that separates structural and numerical prediction. The first model, \textit{QwenA}, predicts where prosodic boundaries occur; the second, \textit{QwenB}, supplies the corresponding numerical attributes.

\subsubsection*{QwenA (Stage 1): Break Prediction}

We fine-tune a Qwen 2.5-7B model (QLoRA with 4-bit quantization, rank 8, $\alpha=16$) to insert \texttt{<break>} tags at linguistically appropriate junctures. QwenA processes up to 200-word French paragraphs (within a 1024-token limit), retaining punctuation, quotations, and parenthetical clauses so that the model must reason over long-range dependencies rather than relying on sentence-level heuristics. These features reflect real-world TTS applications, where systems rarely receive inputs entirely devoid of punctuation or other natural orthographic cues. Furthermore, this approach aligns with the baseline methodology used in \citet{vadapalli2025investigation}. A deterministic post-processor then converts each \texttt{<break>} into an empty \texttt{<prosody>} element, yielding a syntactically valid SSML skeleton to pass into the next stage.

\subsubsection*{QwenB (Stage 2): Prosodic Regression}

QwenB builds on the skeleton emitted by QwenA and replaces each empty \texttt{<prosody>} placeholder with fully specified numeric attributes (pitch, rate, volume, and break duration). Starting again from Qwen 2.5-7B, we inject a second QLoRA adapter with 4-bit quantization (rank 8, $\alpha=16$) into the value and feed-forward projections so that only those low-rank updates are trainable. \textbf{Loss is computed on the numeric tokens}, so categorical text incurs zero penalty and the adapter’s capacity is \textbf{devoted entirely to modeling prosodic distributions}.  Targets are standardized to unit variance during optimization and rescaled at inference, a choice that stabilizes gradients and accelerates convergence.

\section{Results and Analysis}\label{sec:results}

\subsection{Perceptual Evaluation (AB Test)}
\label{subsect:abtest}

To assess our SSML annotation pipeline's effectiveness in enhancing synthetic speech (Section~\ref{sec:dataset}), we conducted AB testing with 18 participants.
Each participant evaluated 30 one-minute audio pairs, where the baseline was the raw, unaltered voice of Microsoft Azure Neural TTS (Henri), without any prosody modifications, compared to the prosody-enhanced version. These pairs were presented randomly, with 60 segments evaluated per participant.

The SSML-enhanced audio achieved a MOS of 3.87 (5-point scale), outperforming the baseline (3.20) and yielding a \textbf{20\% improvement} in perceived quality. Additionally, 15 of 18 participants preferred the enhanced version in over half of the cases, with 7 preferring it in more than 75\% of comparisons. These results support the effectiveness of our SSML-based prosody enhancement for improving synthetic speech quality.

\subsection{Baseline Model Performance}

\paragraph{BERT Break Prediction Results:}

We evaluated the performance of the fine-tuned BERT model from \citet{vadapalli2025investigation} on $F_1$ (\%) and perplexity (best = 1), and attained results very close to the original paper, which reports a 92.10\% $F_1$ score for break prediction. Our model achieves a $F_1$ score of 92.06\%, along with a perplexity of 1.123 (not reported in \citet{vadapalli2025investigation} but useful for further evaluation).
Our stability assessment on 10 bootstrapped datasets yielded an average $F_1$ of 47.52\% $\pm$ 4.65\% (Confidence Interval (CI) = 9.8\%) and perplexity of 1.274 $\pm$ 0.005 (CI = 0.4\%), indicating high stability in token prediction but moderate variability in classification performance. We present the results from the original training data in Table~\ref{tab:break_pred_comparison}.

\paragraph{BiLSTM Prosody Prediction Results:}
We evaluated our BiLSTM model following \citet{pethe2025prosodyanalysisaudiobooks}. Table \ref{tab:bilstm_results} presents the MSE values for normalized z-score prosody features. Our approach achieves SOTA results comparable to those reported in the original paper. For a more comprehensive analysis, we also report the raw RMSE and MAE (\%) for each prosodic parameter.

\begin{table}[htbp!]
    \centering
    \caption{
    BiLSTM-based prosodic attribute prediction across sequence window lengths ($L$).
    Best overall performance is achieved at $L=2$, with lowest MAE for volume and rate, and near-best scores for pitch.
    }
    \label{tab:bilstm_results}
    \setlength{\tabcolsep}{4pt}
    \scalebox{.9}{%
    \begin{tabular}{@{}llccc@{}}
    \toprule
    \textbf{$L$} & \textbf{Metric Type} & \textbf{Pitch} & \textbf{Volume} & \textbf{Rate} \\
    \midrule
    \multirow{3}{*}{1} 
      & Z-score MSE ($\downarrow$)& \cellcolor[gray]{0.9}0.8752 & 0.9141 & 0.7733  \\
      & $\%$ RMSE ($\downarrow$)  & \cellcolor[gray]{0.9}2.0659 & 7.8597 & 1.2771  \\
      & $\%$ MAE ($\downarrow$)   & \cellcolor[gray]{0.9}1.6709 & 6.4768  & 0.8878  \\
    \midrule
    \multirow{3}{*}{2} 
      & Z-score MSE ($\downarrow$)& 0.8983 & \cellcolor[gray]{0.9}0.8949 & \cellcolor[gray]{0.9}0.7572   \\
      & $\%$ RMSE ($\downarrow$) & 2.0930 & \cellcolor[gray]{0.9}7.7767 & \cellcolor[gray]{0.9}1.2637 \\
      & $\%$ MAE ($\downarrow$)   & 1.6883 & \cellcolor[gray]{0.9}6.0405  & \cellcolor[gray]{0.9}0.8462   \\
    \midrule
    \multirow{3}{*}{3} 
      & Z-score MSE ($\downarrow$) & 0.9936 & 0.9917 & 0.8593 \\
      & $\%$ RMSE ($\downarrow$) & 2.2012 & 8.1864 & 1.3462 \\
      & $\%$ MAE ($\downarrow$)   & 1.7732 & 6.5100  & 0.9257  \\
    \midrule
    \multirow{3}{*}{4} 
      & Z-score MSE ($\downarrow$) & 0.9950 & 0.9992 & 0.8263 \\
      & $\%$ RMSE ($\downarrow$) & 2.2028 & 8.2172 & 1.3201 \\
      & $\%$ MAE   ($\downarrow$) & 1.7568 & 6.5990  & 0.9312  \\
    \bottomrule
    \end{tabular}%
    }
    
\end{table}

Unlike \citet{pethe2025prosodyanalysisaudiobooks}, our analysis revealed that a sequence window length ($L=2$) yielded superior performance across prosodic attributes. Specifically, $L=2$ demonstrated lower error rates for two of the three prosodic attributes, while pitch prediction achieved optimal performance with $L=1$. Notably, the MAE for volume with $L=2$ was more than 0.04 percentage points lower than all other tested lengths, and 0.04–0.09 percentage points superior to alternative sequence configurations.
Our best results align with those of \citet{pethe2025prosodyanalysisaudiobooks}: z-scored MSE of 0.8734 for pitch, 0.7631 for volume, and 1.0610 for speaking rate. We attribute minor performance differences to dataset variations and establish the $L=2$ model as our primary baseline for subsequent comparisons with our proposed cascaded architecture.

\subsection{Zero-shot and Few-shot Prompting Evaluation}
\label{subsec:prompting_results}

\begin{figure}[t]
    \centering
    \begin{subfigure}[t]{0.9\columnwidth}
        \centering
        \includegraphics[width=\linewidth]{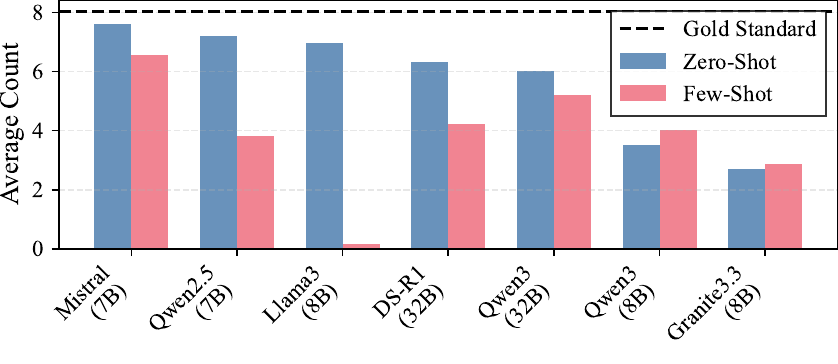}
        \caption{Break tag usage comparison (DS = DeepSeek)}
        \label{fig:break_tags}
    \end{subfigure}
    
    \vspace{0.7em}
    \begin{subfigure}[t]{0.9\columnwidth}
        \centering
        \includegraphics[width=\linewidth]{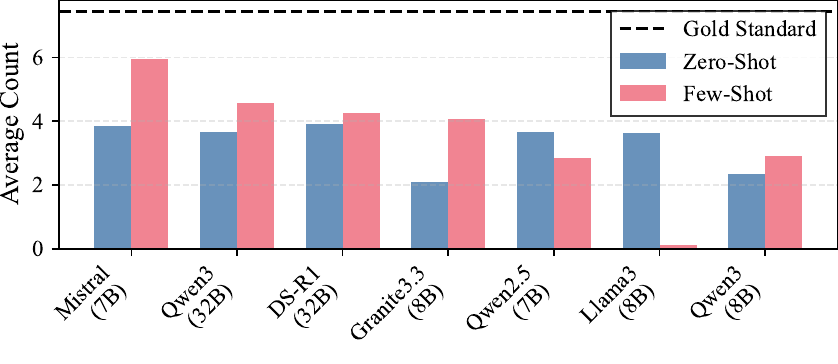}
        \caption{Prosody tag usage comparison (DS = DeepSeek)}
        \label{fig:prosody_tags}
    \end{subfigure}
    
    \caption{Structural comparison of SSML tag predictions across models. All models under-generate both break and prosody tags relative to the gold standard.}
    \label{fig:tags}
\end{figure}

\begin{table*}[htbp!]
\small
\centering
\begin{threeparttable}
\caption{SSML generation performance across models and prompting strategies, evaluated by cosine similarity of predicted vs. gold SSML embeddings, and MAE/\textcolor{blue}{RMSE} for pitch, volume, rate, and break durations. Qwen2.5 (7B) offers the best trade-off between accuracy and efficiency.}
\label{tab:ssml_performance}

\begin{tabular}{l@{\hspace{8pt}}c@{\hspace{8pt}}c@{\hspace{8pt}}c@{\hspace{8pt}}c@{\hspace{8pt}}c@{\hspace{8pt}}c}
\toprule
\textbf{Model} &
\makecell{\textbf{SSML} \\ \textbf{Sim.}\,$\uparrow$} &
\makecell{\textbf{Pitch} (\%) \\ MAE/\textcolor{blue}{RMSE}\,$\downarrow$} &
\makecell{\textbf{Volume} (\%) \\ MAE/\textcolor{blue}{RMSE}\,$\downarrow$} &
\makecell{\textbf{Rate} (\%) \\ MAE/\textcolor{blue}{RMSE}\,$\downarrow$} &
\makecell{\textbf{Break Time} (ms) \\ MAE/\textcolor{blue}{RMSE}\,$\downarrow$} \\
\midrule
Qwen3 (32B) (ZS)          & \cellcolor[gray]{0.9}0.91 & 1.42/\textcolor{blue}{1.83} & 7.65/\textcolor{blue}{8.48}  & 1.52/\textcolor{blue}{2.00}  & 170.23/\textcolor{blue}{232.41} \\
Qwen2.5 (7B) (ZS)         & 0.90                      & 2.07/\textcolor{blue}{2.43} & 7.23/\textcolor{blue}{8.05}  & 1.54/\textcolor{blue}{1.93}  & 361.88/\textcolor{blue}{393.04} \\
Qwen3 (32B) (FS)          & 0.90                      & \cellcolor[gray]{0.9}1.08/\textcolor{blue}{1.41} & 5.80/\textcolor{blue}{7.33}  & 0.97/\textcolor{blue}{1.31}  & 159.58/\textcolor{blue}{215.50} \\
Qwen3 (8B) (FS)           & 0.90                      & 1.77/\textcolor{blue}{2.83} & 6.96/\textcolor{blue}{16.85} & 1.23/\textcolor{blue}{1.69}  & 147.24/\textcolor{blue}{242.98} \\
Qwen2.5 (7B) (FS)         & 0.89                      & 1.26/\textcolor{blue}{1.50} & \cellcolor[gray]{0.9}4.32/\textcolor{blue}{6.77} & 1.01/\textcolor{blue}{1.24}  & \cellcolor[gray]{0.9}118.85/\textcolor{blue}{179.68} \\
Mistral (7B) (ZS)              & 0.88                      & 1.85/\textcolor{blue}{2.25} & 24.19/\textcolor{blue}{43.96} & 18.30/\textcolor{blue}{41.24} & 207.28/\textcolor{blue}{258.76} \\
Mistral (7B) (FS)              & 0.87                      & 1.75/\textcolor{blue}{2.16} & 5.38/\textcolor{blue}{8.33}  & 1.14/\textcolor{blue}{1.42}  & 205.03/\textcolor{blue}{384.17} \\
Granite3.3 (8B) (FS)           & 0.87                      & 1.45/\textcolor{blue}{1.86} & 4.95/\textcolor{blue}{7.12}  & \cellcolor[gray]{0.9}0.95/\textcolor{blue}{1.30} & 196.93/\textcolor{blue}{265.07} \\
Llama3 (8B) (ZS)               & 0.84                      & 1.44/\textcolor{blue}{1.82} & 7.30/\textcolor{blue}{8.08}  & 2.26/\textcolor{blue}{10.17} & 285.17/\textcolor{blue}{318.19} \\
Qwen3 (8B) (ZS)           & 0.82                      & 1.99/\textcolor{blue}{2.70} & 7.43/\textcolor{blue}{8.41}  & 1.69/\textcolor{blue}{2.06}  & 274.27/\textcolor{blue}{334.20} \\
Deepseek-R1 (32B) (ZS)    & 0.81                      & 1.64/\textcolor{blue}{2.11} & 15.50/\textcolor{blue}{30.41} & 18.79/\textcolor{blue}{41.14} & 274.66/\textcolor{blue}{320.62} \\
Granite3.3 (8B) (ZS)           & 0.76                      & 3.70/\textcolor{blue}{4.55} & 13.86/\textcolor{blue}{29.11} & 33.25/\textcolor{blue}{55.85} & 320.77/\textcolor{blue}{413.91} \\
Deepseek-R1 (32B) (FS)    & 0.76                      & 1.43/\textcolor{blue}{2.04} & 7.12/\textcolor{blue}{8.23}  & 3.69/\textcolor{blue}{12.94} & 244.85/\textcolor{blue}{302.87} \\
Llama3 (8B) (FS)               & 0.34                      & 1.26/\textcolor{blue}{1.62} & 7.24/\textcolor{blue}{8.23}  & 1.53/\textcolor{blue}{1.88}  & 416.13/\textcolor{blue}{445.99} \\
\bottomrule
\end{tabular}

\begin{tablenotes}\footnotesize
\item $\uparrow$: higher is better, $\downarrow$: lower is better.  ZS: Zero-Shot, FS: Few-Shot.
\end{tablenotes}
\end{threeparttable}
\end{table*}
We first focused on evaluating \textit{break tag prediction}, a proxy for assessing structural correctness and syntagm segmentation. Figure~\ref{fig:tags} shows the average number of predicted \texttt{<break>} and \texttt{<prosody>} tags per segment compared to gold annotations. 
All models consistently under-generate tags, indicating systematic issues maintaining SSML structure. Few-shot prompting led to unexpected patterns: fewer predicted break tags but increased \texttt{<prosody>} tags, suggesting attention shifts or stylistic overfitting to prompt exemplars. 
Notably, Llama 3 and DeepSeek-R1 (32B) show large discrepancies between zero- and few-shot modes, with Llama 3’s prosody tagging almost collapsing in the few-shot case.

Beyond structural accuracy, we evaluated numerical performance through \textit{cosine similarity} between predicted and reference SSML structures, embedded using the \textit{all-MiniLM-L6-v2} \footnote{\url{https://huggingface.co/sentence-transformers/all-MiniLM-L6-v2}} model, RMSE, and MAE for break durations and prosodic coefficients, averaged per segment. Table~\ref{tab:ssml_performance} summarizes the results. 
Qwen 2.5-7B achieves the best overall balance: in the few-shot setting, it delivers the lowest MAE for break (118.85 ms) and volume (4.32\%), and second-highest structural similarity in zero-shot (0.9). Qwen 3 (32B) slightly surpasses it on similarity (0.908), but at a cost of 4.5 times higher memory usage and slower inference, making it less suitable for fine-tuning and deployment.

Our findings suggest that while few-shot prompting can improve prosody tag usage and numerical accuracy, model behavior is highly architecture-dependent. Furthermore, the consistent underproduction of tags across models highlights the need for more robust SSML-structure awareness.

\subsection{Cascaded LLM Evaluation}
\label{subsec:cascaded_evaluation}

Our evaluation of the cascaded QwenA and QwenB models demonstrates substantial performance improvements over existing SOTA approaches, as detailed in Table~\ref{tab:break_pred_comparison}:

\begin{table}[htbp!]
  \centering
  \caption{Break tag prediction: F1 and perplexity for our cascaded model (QwenA) vs.\ fine-tuned BERT. QwenA achieves near-perfect accuracy and fluency.}
  \label{tab:break_pred_comparison}
  \setlength{\tabcolsep}{2pt} % Reduced column separation
  \resizebox{1\linewidth}{!}{%
    \begin{tabular}{@{}lcc@{}}
      \toprule
      \textbf{Model} & \textbf{F1} (\%) & \textbf{Perplexity} ($\rightarrow\,1$) \\
      \midrule
      Cascade (QwenA) & \cellcolor[gray]{0.9}99.24 & \cellcolor[gray]{0.9}1.00 \\
      Finetuned BERT   & 92.06 & 1.12 \\
      \bottomrule
    \end{tabular}%
  }% 
\end{table}

For QwenA, which utilizes next-token prediction on a linearized SSML target, the model achieved a test \textbf{perplexity of 1.001 and a tag-level $F_1$ score of 99.24\%}, surpassing the fine-tuned BERT's baseline of 1.123 perplexity and 92.06\% $F_1$ score. Moreover, this approach also outperforms the LLM tag prediction benchmarks, which consistently under-generate break and prosody tags, as illustrated in Figures~\ref{fig:break_tags} and \ref{fig:prosody_tags}. This near-perfect tag insertion accuracy validates the improved performance of our cascaded approach compared to available models for SSML tag prediction.

\begin{table}[t]

    \caption{RMSE ($\downarrow$) and MAE ($\downarrow$) for our cascaded model vs.\ benchmarks. It achieves the lowest error scores across nearly all prosody attributes.}
    \label{tab:evaluation_results}
    \setlength{\tabcolsep}{2pt}
    \scalebox{0.90}{%
    \begin{tabular}{@{}llcccc@{}}
    \toprule
    \textbf{Model} & \textbf{Metric} & \textbf{Pitch} & \textbf{Volume} & \textbf{Rate} & \textbf{Break Time} \\
    \midrule
        %-------------------- Cascade --------------------
    \multirow{2}{*}{\shortstack{Cascade\\ (Ours)}} & RMSE & \cellcolor[gray]{0.9}1.22 & \cellcolor[gray]{0.9}1.67 & 1.50 & \cellcolor[gray]{0.9}166.51 \\
                             & MAE & \cellcolor[gray]{0.9}0.97 & \cellcolor[gray]{0.9}1.09 & 1.10 & \cellcolor[gray]{0.9}132.89 \\
    \midrule
    %-------------------- Bi-LSTM --------------------
    \multirow{2}{*}{\shortstack{BiLSTM$^\dag$ \\ ($L=2$)}} & RMSE & 2.09 & 7.77 & \cellcolor[gray]{0.9}1.26 & -- \\
                             & MAE & 1.68 & 6.04  & \cellcolor[gray]{0.9}0.84 & -- \\
    \midrule
    %-------------------- SOTA Few-Shot ---------------
    \multirow{2}{*}{\shortstack{SOTA \\ Few-Shot*}} & RMSE & 1.41 & 7.33 & 1.31 & 215.50 \\
                                                    & MAE & 1.08 & 5.80  & 0.97 & 159.58 \\
    \midrule
    %-------------------- SOTA Zero-Shot --------------
    \multirow{2}{*}{\shortstack{SOTA \\ Zero-Shot*}} & RMSE & 1.83 & 8.48 & 2.00 & 232.41 \\

                                                     & MAE & 1.42 & 7.65  & 1.52 & 170.23 \\
    \midrule
    \multicolumn{6}{@{}l}{$\dag$: Results based on \citet{pethe2025prosodyanalysisaudiobooks}; see Section~\ref{subsec:bilstm_baseline}} \\
    \multicolumn{6}{@{}l}{$*$: Qwen-3 (32B) selected via cosine similarity (Tab.~\ref{tab:ssml_performance})} \\
    \multicolumn{6}{@{}l}{Units: Pitch, Volume, Rate (\%); Break Time (ms)} \\
    \bottomrule
    \end{tabular}%
    }
\end{table}

QwenB demonstrates significant advancements in prosody parameter prediction, achieving an \textbf{MAE of 0.97\% for pitch, 1.09\% for volume, 1.10\% for rate, and 132.9ms for break timing} (Table~\ref{tab:evaluation_results}). Furthermore, this strong performance is achieved while maintaining an efficient end-to-end latency of approximately 190~ms for a 150-word paragraph. This demonstrates the model's enhanced SSML parameter prediction and its ability to process larger text segments, outperforming baseline approaches. This performance also suggests that evaluations of pipeline audio (Section~\ref{subsect:abtest}) are highly generalizable to the cascaded model's audio quality due to their close similarity.

\subsection{Summary and Analysis of Results}
\label{subsec:results_discussion}

Table~\ref{tab:evaluation_results} provides a comparative overview of objective performance across all evaluated approaches, revealing three key observations:
\begin{enumerate}
    \item \textbf{Cascaded QwenA + QwenB sets new SOTA performance.} The system achieves single-digit MAE for all prosodic coefficients and reduces break-timing error by 25\% vs. the best few-shot LLM baseline. 

    \item \textbf{BiLSTM architectures remain competitive for speaking rate prediction.} 
    Though outperformed elsewhere, its 0.84\% MAE on rate shows lightweight sequential models still capture localized prosodic patterns effectively.
    \item \textbf{Prompt-only LLMs systematically under-generate tags.} Both zero- and few-shot settings underperform supervised baselines on break timing prediction (MAE > 150~ms) and structural metrics (Figure~\ref{fig:tags}), reinforcing the necessity for explicit structural supervision in SSML generation tasks.
\end{enumerate}

These findings confirm that disentangling structural prediction (QwenA) from numerical regression (QwenB) yields optimal performance across both dimensions: syntactically valid SSML markup with fine-grained prosodic control, while preserving inference efficiency suitable for real-time TTS applications. The subjective evaluation results in Section~\ref{subsect:abtest} corroborate these objective improvements, demonstrating that enhanced technical performance translates into substantial perceptual gains—a 20\% MOS improvement and consistent listener preference for enhanced synthesis.

\section{Conclusion and Future Directions}
\label{sec:conclusion} 
Using a fine-tuned cascaded Qwen 2.5-7B architecture, we separate structural tag insertion from prosodic parameter prediction, achieving near-perfect break placement (99.2\% $F_1$, perplexity 1.001) and reducing prosodic MAE below 1.1 points -- representing 25--40\% better than prompting-only LLMs and BiLSTM baselines.

Perceptual evaluation shows that SSML from our pipeline increases MOS from 3.20 to 3.87, with consistent listener preference. This marks a significant step toward closing the expressiveness gap between synthetic and natural French speech while preserving compatibility with commercial TTS.

Future research includes unifying our cascaded approach into a single end-to-end model for joint prosodic prediction, incorporating multimodal audio embeddings to capture subtle speech characteristics beyond text-derived features, and extending this methodology to additional languages to assess cross-linguistic generalizability and robustness.

\clearpage
\section{Limitations}
While our proposed system shows significant improvements, several limitations warrant discussion. Our experiments focus exclusively on French using a proprietary 14-hour corpus. While our pipeline is language-agnostic, performance may vary for languages with different prosodic characteristics. The dataset size remains modest compared to typical TTS training corpora, as English prosody modeling often leverages hundreds of hours of annotated speech, indicating that scaling our French dataset could yield additional performance gains. Additionally, our improvements rely on TTS engines supporting fine-grained SSML tags, meaning legacy or non-compliant systems may not achieve similar gains and may require custom adjustments for engine-specific behaviors.

Our prosodic deltas are computed with respect to a single baseline synthetic voice (Azure fr-FR-HenriNeural) and evaluated with the same voice, which limits out-of-domain generalization. While SSML prosody tags are standardized, their acoustic realization is implementation- and voice-dependent; engines may clamp or substitute values, and different voices can map the same percentage to different F0/rate changes. Consequently, numeric SSML settings may require voice-specific recalibration (e.g., a short script that sweeps pitch/rate/volume and measures resulting semitone, syllables/s, and dB changes) before transfer to other voices or engines.

From a computational perspective, fine-tuning Qwen 2.5-7B requires substantial GPU memory ($\approx 15~GB$ peak) despite 4-bit quantization, necessitating model compression or distillation for smaller deployments. Conversely, greater computational resources could enable more extensive fine-tuning and potentially improve performance. Our approach also assumes that punctuation and syntactic cues correlate well with natural prosodic boundaries, an assumption that may break down in highly informal or unpunctuated text such as social media transcripts, leading to suboptimal break placement.

\section{Ethics Statement}
Our work uses commercially licensed French podcast audio, ensuring no personal or sensitive data are exposed. We acknowledge potential biases from using a limited speaker set and encourage broader demographic validation. While improved prosody can enhance synthetic voices, it also risks misuse in deceptive audio generation; we therefore recommend watermarking or verification mechanisms. Code and anonymized alignment scripts are publicly shared to promote reproducibility and transparency.

\section*{Acknowledgments}
This work was supported by Hi! PARIS and by the ANR/France 2030 program (ANR-23-IACL-0005).\\
We acknowledge access to the IDRIS high-performance computing resources under allocation \textit{20XX-AD011015141R2}, granted by GENCI (Grand Équipement National de Calcul Intensif).\\
\textit{Ce projet a été financé par l’État dans le cadre de France 2030.}\\
\textit{Financé par l’Union européenne – NextGenerationEU dans le cadre du plan France Relance.}

\newpage
\appendix
\label{sec:appendix}
\section{Dataset Statistics}
\label{sec:appendix_dataset}

The dataset constructed through our end-to-end SSML annotation pipeline (Section~\ref{sec:dataset}) comprises 14 speakers (42\% female), encompassing 122,303 words across 711,603 characters. Our annotation process generated 17,695 \texttt{<prosody>} tags and 18,746 \texttt{<break>} tags, providing comprehensive prosodic markup for the corpus (Table~\ref{tab:corpus_stats}).

\begin{table}[htbp]
\centering
\caption{Corpus statistics for the annotated French speech dataset}
\label{tab:corpus_stats}
\begin{tabular}{lr}
\toprule
\textbf{Metric} & \textbf{Value} \\
\midrule
Speakers & 14 \\
Total characters & 711,603 \\
Total words & 122,303 \\
Prosody tags & 17,695 \\
Break tags & 18,746 \\
\bottomrule
\end{tabular}
\end{table}

\begin{figure}[htbp]
    \centering
    \includegraphics[width=0.9\linewidth]{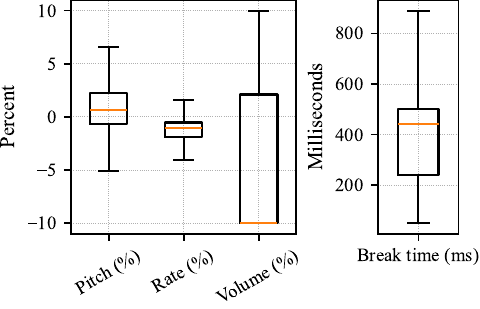}
    \caption{Distribution of prosodic parameters in the annotated dataset. \textbf{Left:} Pitch, rate, and volume adjustments (percentage) relative to synthetic baseline. \textbf{Right:} Break durations (milliseconds) derived from natural inter-phrasal pauses.}
    \label{fig:dataset_distributions}
\end{figure}

The prosodic parameter distributions reveal linguistically meaningful patterns (Figure~\ref{fig:dataset_distributions}). Pitch adjustments cluster around +1\% with 50\% of values within $\pm$2\%, reflecting the subtle phrase-final rises characteristic of French declarative intonation. Rate modifications center at -1\%, indicating slight deceleration relative to the neutral Azure baseline, consistent with the deliberate pacing typical of podcast narration. Volume adjustments concentrate at -10\% with an upper bound at +2\%, reflecting our systematic reduction strategy relative to the synthetic baseline to achieve more natural amplitude levels.

Break duration analysis reveals a median pause of approximately 400~ms with an interquartile range of 250--500~ms, aligning with established phonetic studies on French prosodic phrase boundaries~\cite{peck_syntactic_2024,campione_large-scale_2002}.

\section{Comparative Analysis of Prosodic Features}
\label{sec:appendix_prosodic_analysis}

\subsection{Pitch Characteristics}

\begin{figure}[htbp]
    \centering  
    \includegraphics[width=\linewidth]{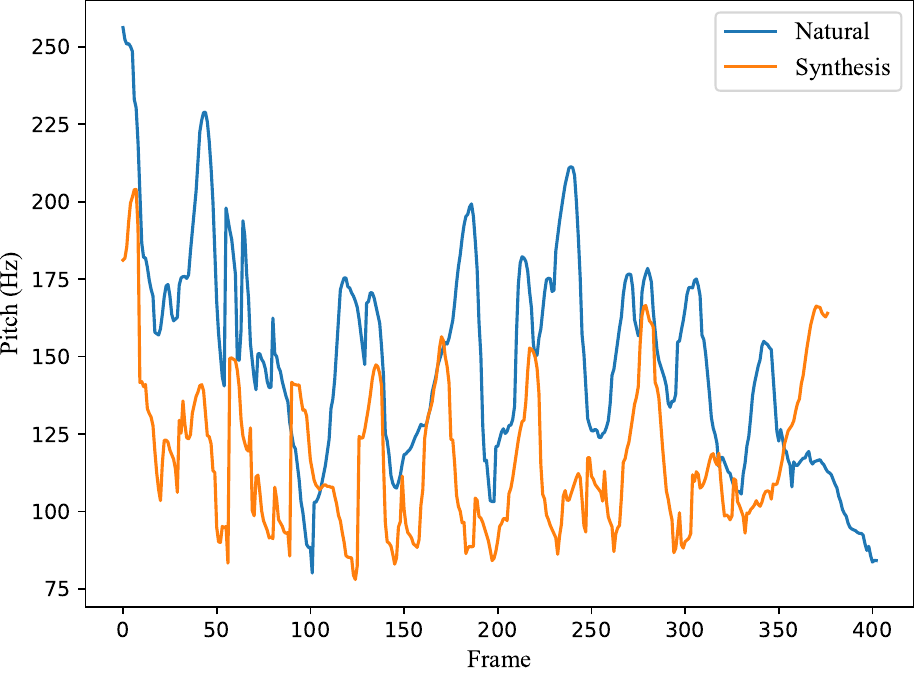}
    \caption{Temporal pitch contours, comparing natural and synthesized speech across representative utterances}
    \label{fig:pitch_temporal}
\end{figure}

Figure~\ref{fig:pitch_temporal} demonstrates the temporal evolution of fundamental frequency in natural versus synthesized speech. Natural speech exhibits a broader pitch range with complex, fluid intonational patterns reflecting the dynamic modulation inherent in human vocal production. Conversely, synthesized speech operates within a constrained, generally lower fundamental frequency range, displaying more abrupt transitions and reduced prosodic variability.

\begin{figure}[htbp]
    \centering
    \includegraphics[width=\linewidth]{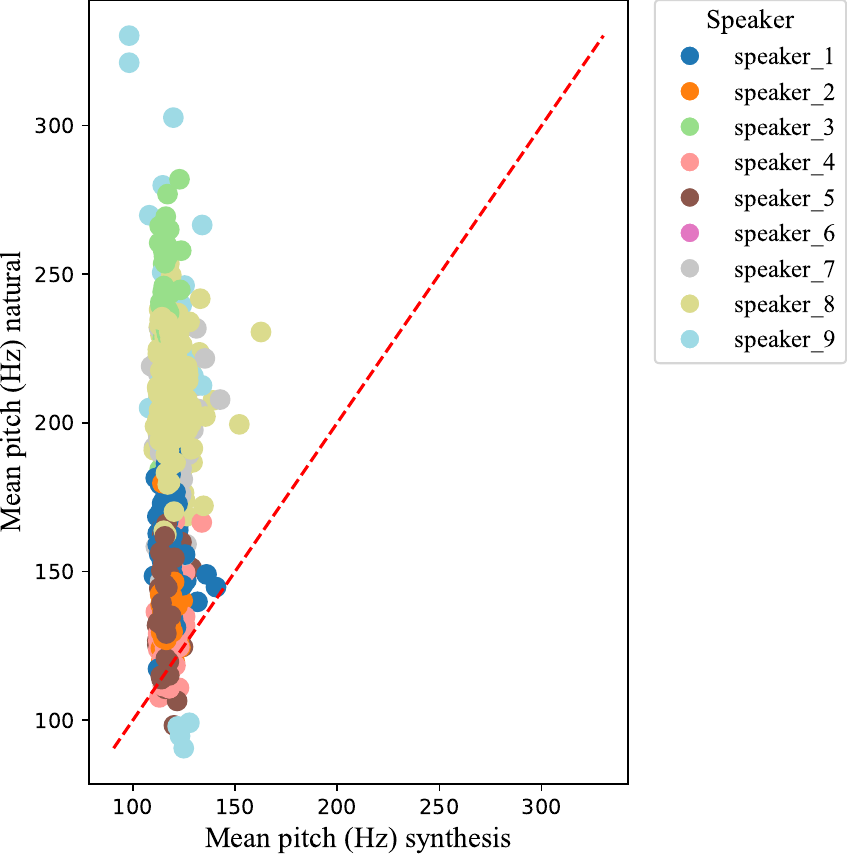}
    \caption{Speaker-wise mean pitch comparison: natural speech (y-axis) versus synthesized speech (x-axis). Each point represents one speaker's average fundamental frequency.}
    \label{fig:pitch_scatter}
\end{figure}

Cross-speaker analysis (Figure~\ref{fig:pitch_scatter}) reveals substantial inter-speaker pitch variability in natural speech, while synthesized versions cluster within a significantly narrower frequency range. This compression of the pitch space in synthetic speech represents a fundamental limitation in current TTS systems' ability to capture individual vocal characteristics.

\subsection{Volume Dynamics}

\begin{figure}[htbp]
    \centering
    \includegraphics[width=\linewidth]{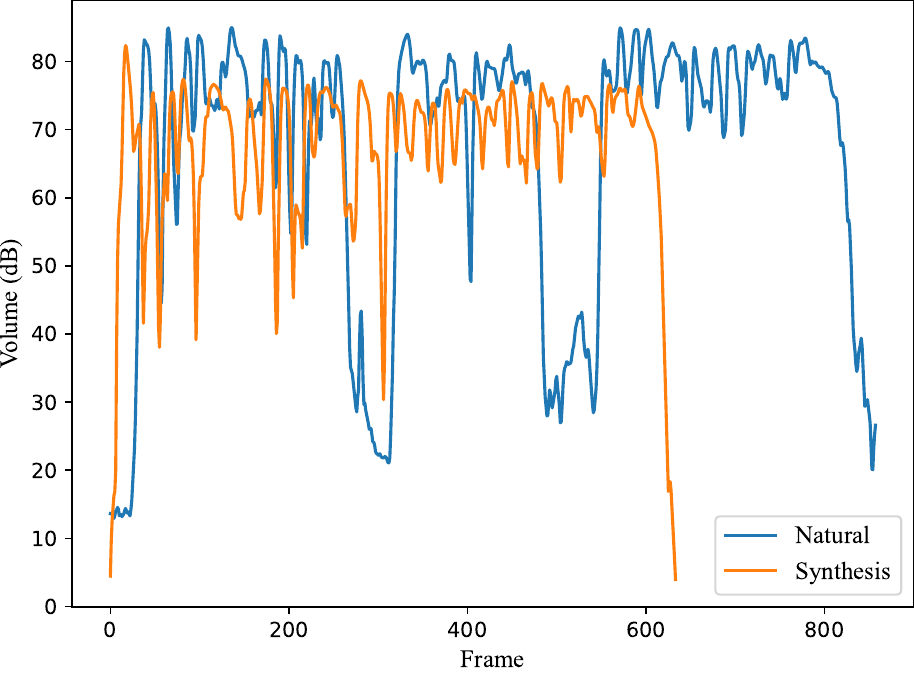}
    \caption{Volume variation patterns over time for natural versus synthesized speech}
    \label{fig:volume_temporal}
\end{figure}

Amplitude modulation patterns (Figure~\ref{fig:volume_temporal}) reveal marked differences between natural and synthetic speech production. Natural speech demonstrates substantial dynamic range with frequent amplitude variations, characteristic of expressive human discourse and reflecting the speaker's communicative intent. Synthesized speech exhibits limited volume variation, maintaining relatively consistent amplitude levels that contribute to reduced prosodic expressiveness.

\begin{figure}[htbp]
    \centering
    \includegraphics[width=\linewidth]{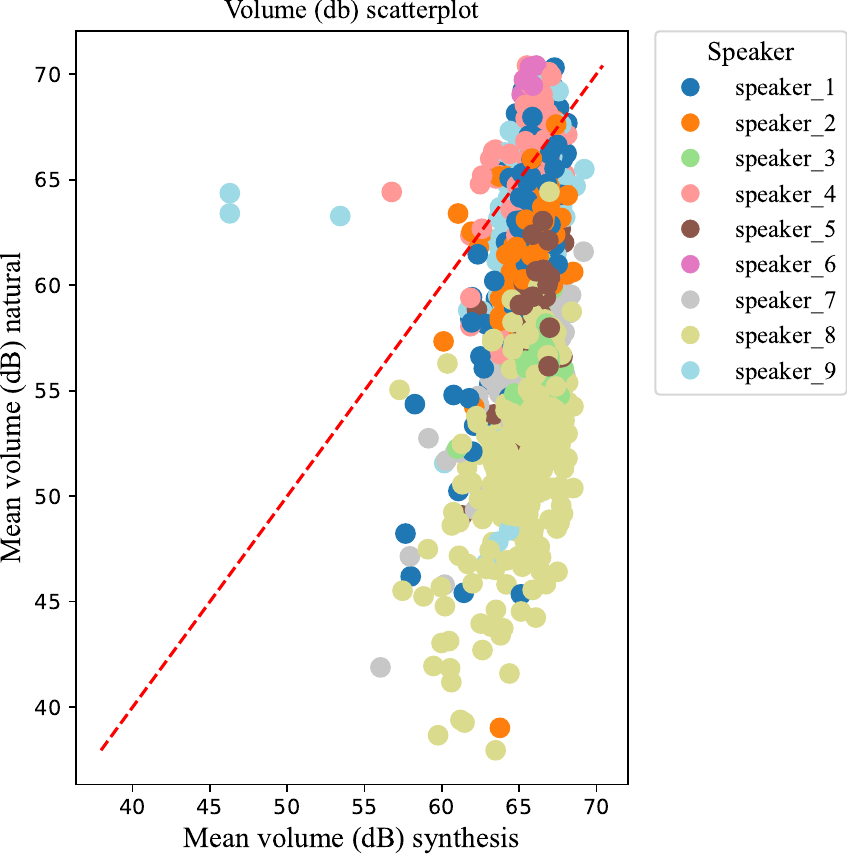}
    \caption{Speaker-wise mean volume comparison between natural and synthesized speech}
    \label{fig:volume_scatter}
\end{figure}

Speaker-level volume analysis (Figure~\ref{fig:volume_scatter}) confirms the systematic amplitude differences between natural and synthetic speech across all speakers in our corpus.

\section{Evaluation Metrics}
\label{sec:appendix_metrics}

Our evaluation employs standard, well-established metrics from the speech processing and natural language processing domains:
\begin{align}
\text{Perplexity} &= \exp\!\bigl(\text{CrossEntropy}(p,q)\bigr), \\
F_1 &= 2 \cdot \frac{\text{Precision} \times \text{Recall}}{\text{Precision} + \text{Recall}}, \\
\text{WER} &= \frac{S + D + I}{N}, \\
\text{MAE} &= \frac{1}{n} \sum_{i=1}^{n} |P_i - A_i|, \\
\text{RMSE} &= \sqrt{\frac{1}{n} \sum_{i=1}^{n} (P_i - A_i)^2}, \\
\text{ARR} &= \frac{\bigl|\{\,\text{words aligned within } \tau\,\}\bigr|}{N}.
\end{align}

\noindent Here, $p$ and $q$ denote the true and predicted distributions (perplexity). In WER, $S$, $D$, and $I$ are substitutions, deletions, and insertions, and $N$ is the number of reference words. In MAE and RMSE, $n$ is the number of predictions, with $P_i$ and $A_i$ the predicted and actual values for instance $i$. For ARR (Alignment Recall Rate), $\tau$ is the temporal tolerance for correct alignment (e.g., $\pm 50$~ms). Unless otherwise specified, we report a macro-averaged ARR: the ratio is computed in each 15-second window and then averaged over all windows.

\section{SSML Annotation Example}
\label{sec:appendix_ssml_example}

\begin{figure}[htbp]
    \centering
    \includegraphics[width=1\linewidth]{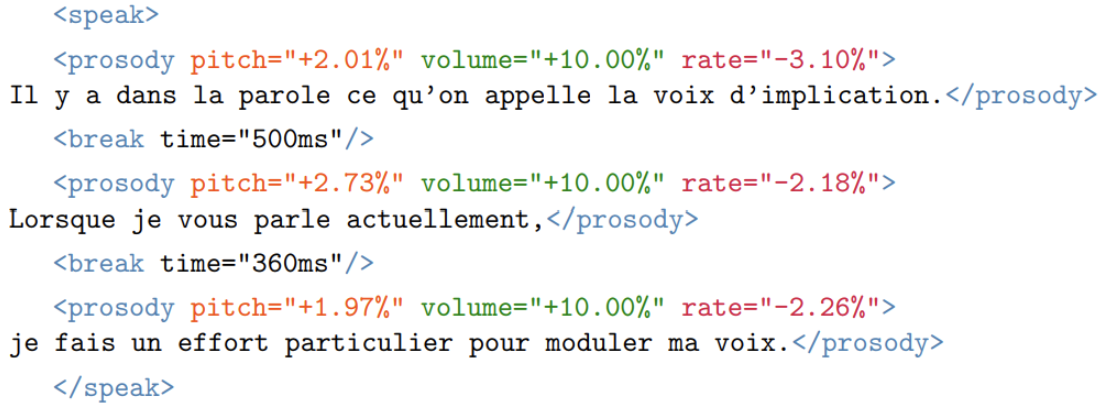}
    \caption{Example of text annotated with SSML prosodic tags generated by our pipeline}
    \label{fig:ssml_example}
\end{figure}

Figure~\ref{fig:ssml_example} illustrates a representative example of our automated SSML annotation, demonstrating the integration of prosodic tags with natural text to enable fine-grained control over synthetic speech parameters.

\end{document}